\documentclass[letterpaper]{article} 
\usepackage{aaai24}  
\usepackage{times}  
\usepackage{helvet}  
\usepackage{courier}  
\usepackage[hyphens]{url}  
\usepackage{graphicx} 
\usepackage{natbib}  
\usepackage{caption} 
\usepackage{algorithm}
\usepackage{algorithmic}
\usepackage{multirow}
\usepackage{newfloat}
\usepackage{listings}
\usepackage{amssymb}
\usepackage{amsmath}
\DeclareCaptionStyle{ruled}{labelfont=normalfont,labelsep=colon,strut=off} 
\floatstyle{ruled}
\newfloat{listing}{tb}{lst}{}
\floatname{listing}{Listing}
\title{Prompting Segmentation with Sound Is \\ Generalizable Audio-Visual Source Localizer}

\author{
Yaoting Wang\textsuperscript{\rm 1}\equalcontrib, 
Weisong Liu\textsuperscript{\rm 2}\equalcontrib, 
Guangyao Li\textsuperscript{\rm 1}, 
Jian Ding\textsuperscript{\rm 3}, 
Di Hu\textsuperscript{\rm 1}\thanks{Corresponding author}, 
Xi Li\textsuperscript{\rm 4}
}

\affiliations{
    \textsuperscript{\rm 1} Gaoling School of Artificial Intelligence, Renmin University of China\\
    \textsuperscript{\rm 2} School of Computer Science, Northwest Polytechnical University\\ 
    \textsuperscript{\rm 3} LIESMARS, Wuhan University\\ 
    \textsuperscript{\rm 4} College of Computer Science and Technology, Zhejiang University\\
    yaoting.wang@outlook.com, liuweisong@mail.nwpu.edu.cn, guangyaoli@ruc.edu.cn\\ jian.ding@whu.edu.cn, dihu@ruc.edu.cn, xilizju@zju.edu.cn
}

\begin{document}

\maketitle

\begin{abstract}
Never having seen an object and heard its sound simultaneously, can the model still accurately localize its visual position from the input audio? In this work, we concentrate on the Audio-Visual Localization and Segmentation tasks but under the demanding zero-shot and few-shot scenarios. To achieve this goal, different from existing approaches that mostly employ the encoder-fusion-decoder paradigm to decode localization information from the fused audio-visual feature, we introduce the encoder-prompt-decoder paradigm, aiming to better fit the data scarcity and varying data distribution dilemmas with the help of abundant knowledge from pre-trained models. Specifically, we first propose to construct a Semantic-aware Audio Prompt (SAP) to help the visual foundation model focus on sounding objects, meanwhile, the semantic gap between the visual and audio modalities is also encouraged to shrink. Then, we develop a Correlation Adapter (ColA) to keep minimal training efforts as well as maintain adequate knowledge of the visual foundation model. By equipping with these means, extensive experiments demonstrate that this new paradigm outperforms other fusion-based methods in both the unseen class and cross-dataset settings. We hope that our work can further promote the generalization study of Audio-Visual Localization and Segmentation in practical application scenarios. Project page: https://github.com/GeWu-Lab/Generalizable-Audio-Visual-Segmentation
\end{abstract}

\label{introduction}

\begin{figure*}[!htbp]
     \centering     
     \includegraphics[width=0.95\textwidth]{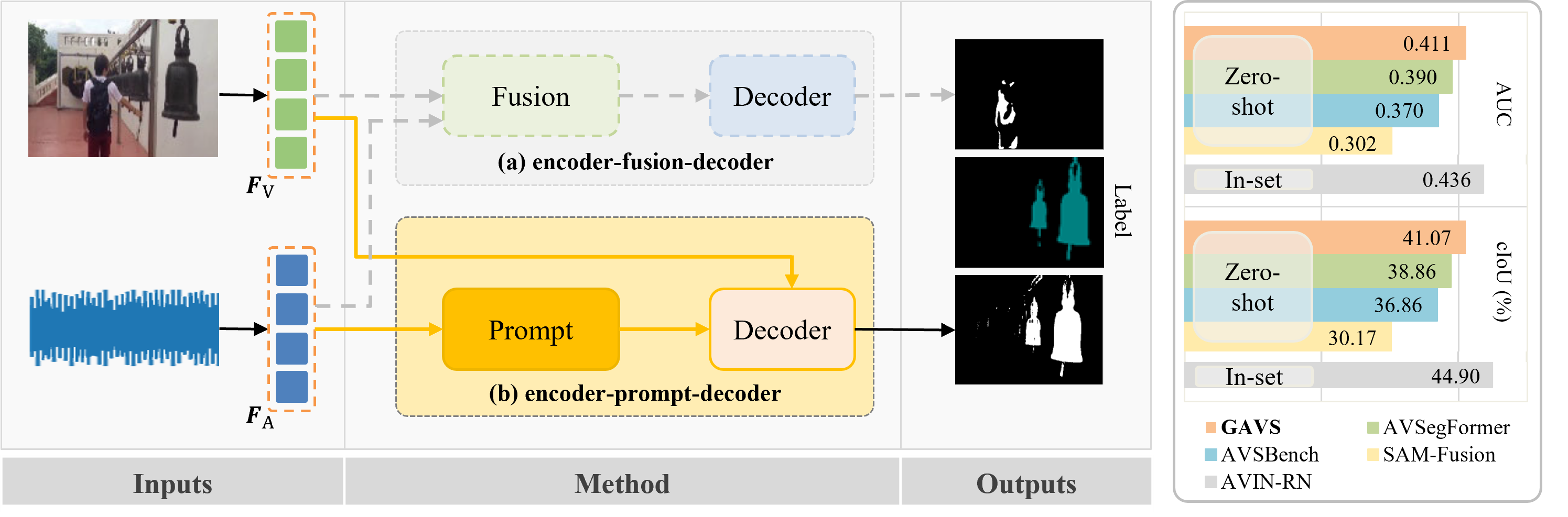}
     \vspace{-0.65em}
     \caption{The AVS pipeline of encoder-fusion-decoder (the upper-center) and our proposed encoder-prompt-decoder (the lower-center) paradigms. Classical encoder-fusion-decoder methods decode mask from the fused modality while we prompting visual input with audio to adapt AVL and AVS tasks to the visual foundational model. The results on the VGG-SS dataset highlight the challenge of generalizing across different datasets. However, our approach breaks through the 40\% cIoU barrier, getting the performance closer to the best trained on in-set (VGG-Sound) method.}
     \label{fig:teaser}
     \vspace{-0.75em}
\end{figure*}

\section{Introduction}
Audio-Visual Localization (AVL) aims to locate the region of sounding objects within a visual scene based on the input audio \cite{wei2022learning, arandjelovic2018objects, chen2021localizing}, which has made satisfactory development with the help of multimodal learning \cite{baltruvsaitis2018multimodal} in recent years. Moreover, the demand for more precise localization in real-world scenarios has led the task of AVL to shift from localization with bounding box or coarse heatmap to finer pixel-level segmentation masks, i.e., Audio-Visual Segmentation (AVS) \cite{zhou2022audio}. 

Currently, as illustrated in the upper-center of Figure \ref{fig:teaser}, many methods commonly implement AVS based on cross-modal correlation learning, such as calculating audio-visual heatmaps through representation similarity \cite{arandjelovic2018objects} or attention \cite{zhou2022audio}. These methods initially fuse the latent features of audio and visual modalities, then learn to localize sounding objects with the fused representation. We name such methods as the encoder-fusion-decoder paradigm. However, in real-world applications, the limited training data and varying data distribution hinder the segmentation performance of models when faced with unseen classes and different datasets (cross-datasets). Therefore, this study focuses on generalizable audio-visual segmentation, which facilitates effective localization for both unseen classes and cross-dataset settings.

To probe the generalization capability of the current encoder-fusion-decoder paradigm, we conduct cross-dataset tests on the VGG-SS dataset~\cite{chen2021localizing} but trained on AVS-Benchmarks~\cite{zhou2022audio}. The right side of Figure \ref{fig:teaser} demonstrates that fusion-based models under the zero-shot setting are unable to surpass the performance of the classic AVL models trained on the VGG-Sound dataset \cite{chen2020vggsound}, which has the same data distribution with VGG-SS. We attribute this performance to the limited generalization ability resulting from exploring audio-visual correlations on specific datasets using the encoder-fusion-decoder paradigm, without prior knowledge from pre-trained models. SLAVC~\cite{mo2022closer} demonstrates that leveraging the prior knowledge within pre-trained visual models can improve the generalization ability. 

We argue that one of the ways to enhance generalization capability is to leverage the prior knowledge encoded in large-scale pre-trained models \cite{yang2023foundation}. Many models in natural language processing (NLP) and computer vision (CV) exhibit remarkable generalization abilities \cite{brown2020language, he2016deep}. Some researchers \cite{li2022learning, zheng2022prompt, zang2022unified} consider prompt learning to enhance the model generalization ability. One of the key benefits lies in its ability to align the data distribution of downstream tasks with the prior knowledge embedded in the foundation model, as the task formats and the output space have reached a consensus between pre-trained models and downstream tasks \cite{shu2022test, jia2022prompt}, consequently enhancing the model's generalization capability across diverse downstream tasks. Drawing inspiration from prompt learning in NLP and multimodal research, we consider that a visual foundation model incorporating audio context cues holds great promise for achieving generalizable AVL and AVS.

Therefore, we introduce an encoder-prompt-decoder paradigm that instructs the visual foundation model to perform sounding object segmentation using audio cues, rather than solely decoding from the fused modality. This paradigm facilitates the seamless integration of the AVS task within the underlying visual foundation model, thereby enhancing the generalization capability of prompt-based models in AVL and AVS through effective utilization of the pre-trained model's prior knowledge. Firstly, we construct a Semantic-aware Audio Prompt (SAP) to bridge the semantic gap between the visual and auditory modalities, aligning the semantics of the given image and audio through contrastive learning. SAP assists the visual foundation model in localizing objects based on the provided audio cues with the same cross-modal semantics. Subsequently, we use a correlation adapter (ColA) to construct the audio-visual correlation to retain as much prior knowledge as possible from the visual foundation model. We use the Segment Anything Model (SAM) \cite{kirillov2023segment} as our visual foundation model for its remarkable segmentation capabilities in generalization-sensitive scenarios.

To evaluate the effectiveness of our method, we first verify the segmentation performance of our Generalizable Audio-visual Segmentation (GAVS) method on AVS-Benchmarks \cite{zhou2022audio}, then we evaluate the zero-shot and few-shot generalization capabilities on AVS-V3 and VGG-SS \cite{chen2021localizing} for unseen classes and cross-dataset settings respectively. Experimental results demonstrate that our model achieves superior generalizable segmentation performance and outstanding few-shot learning ability compared to fusion-based models. In summary, our contributions can be summarized as follows:
\begin{itemize}
\item We investigate the under-explored generalization issue in the AVS task and introduce an encoder-prompt-decoder paradigm to enhance the generalization of the AVS model by leveraging the prior knowledge of the visual foundation model.

\item We introduce a Semantic-aware Audio Prompt (SAP) to assist the model in focusing on the regions of the image that share the same semantics as the given audio. 

\item We propose a Correlation Adapter (ColA) to construct the audio-visual correlation but retain the prior knowledge of the visual foundation model. 
\end{itemize}

\begin{figure*}[!ht]
     \centering
     \includegraphics[width=0.935\textwidth]{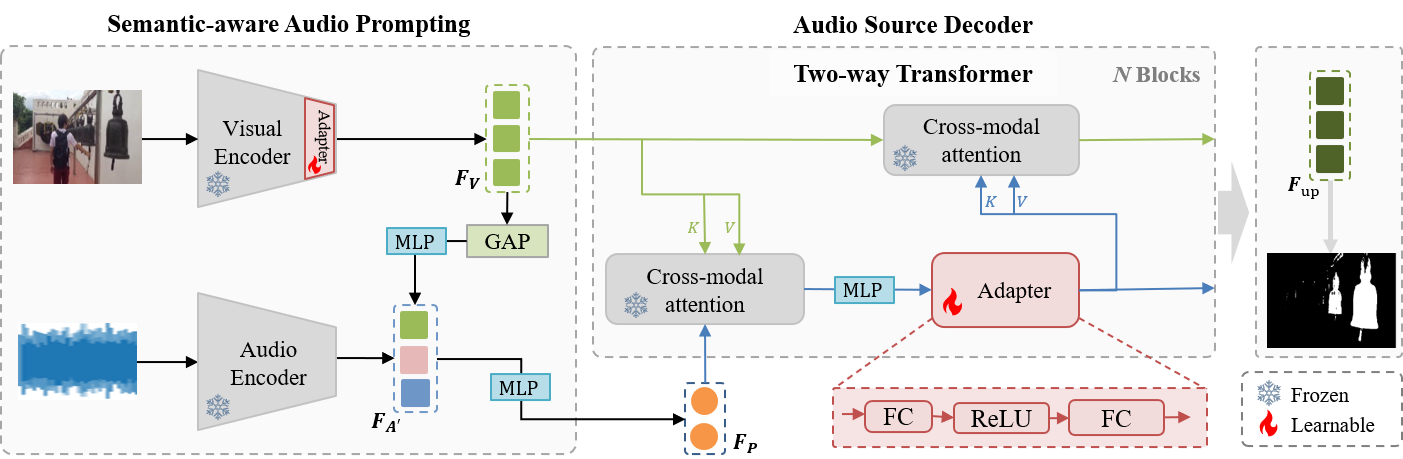}
     \vspace{-0.5em}
     \caption{The overview of GAVS. (1)  We firstly align the audio and visual semantics for SAP, and introduce visual features as cues (the green one in $F_{A^\prime}$) for audio input (the blue one in $F_{A^\prime}$). Then we further combine audio input with learnable adaptive noise (the pink one in $F_{A^\prime}$) to construct the final SAP $F_{A^\prime}$, and get the projected prompt $F_P$. (2) Next, we utilize cross-modal attention to learn the correlation between audio and visual in the Audio Source Decoder, projecting audio into the visual space. The self-attention for $F_P$ before the first cross-modal attention is omitted for clarity.}
     \label{fig:model}
     \vspace{-0.75em}
\end{figure*}

\section{Related Work}
\label{relatedwork}
\subsection{Audio-Visual Localization and Segmentation}
AVL aims to predict the location of sounding objects in a video \cite{wei2022learning}. The traditional AVL task \cite{senocak2018learning, hu2021class, chen2021localizing, mo2022closer, mo2022localizing, park2023marginnce} is typically unsupervised, where the goal is to predict the bounding box or coarse heatmap of the object's location by jointly learning the correspondences between audio and visual features. In recent years, AVL studies have gradually shifted towards learning audio-visual correspondence through the contrastive learning of positive and negative examples. 

A more challenging task of sound source localization, AVS \cite{zhou2022audio}, has been proposed recently, which is a complex extension of the AVL task, as it requires a pixel-level shape description besides localization. AVSBench \cite{zhou2022audio} utilizes multi-stage audio-visual feature fusion to perform a supervised segmentation task on a midsize dataset, predicting the probability of each pixel in the image belonging to the sounding object. AuTR \cite{liu2023audioaware} proposes an audio-aware query-enhanced transformer to address the limitations of small receptive fields in convolutions and inadequate fusion of audio-visual features. AVSC \cite{liu2023audiovisual} presents an audio-visual instance-aware approach to address the ambiguity of silent objects and explore audio-visual semantic correlation to highlight corresponding sounding instances. AUSS \cite{ling2023hear} proposes to unmix complicated audio signals and distinguish similar sounds. AVSegFormer \cite{gao2023avsegformer} employs the transformer architecture to decode fused audio-visual features and utilize audio queries to enhance the model's focus on sounding objects in the visual space. AV-SAM \cite{mo2023av} leverages the promptable nature of SAM to accomplish AVS, however, it still employs fused modalities at the pixel level as the prompt input. This approach of prompt constructing still fails to avoid the problem of insufficient information prompts caused by data scarcity and diverse data distributions. In the experimental section, we further assess the feasibility of AV-SAM's encoder-fusion-prompt-decoder paradigm by implementing a simple Audio-SAM model.

Overall, current research on AVS are primarily focused on close-set and in-domain situations and obtained satisfactory results to some extent. However, there has been a lack of emphasis on investigating the generalization ability in unseen classes and varying data distribution scenarios.

\subsection{Prompt Learning}
Most pre-trained language models are trained using language modelling objectives, which may differ significantly from the objectives of downstream tasks. Consequently, several studies \cite{li2021prefix, lester2021power, liu2023pre} have introduced prompt learning to bridge the gap between pre-training and downstream tasks. In essence, prompt learning assists the model's learning process during pre-training by providing task-specific cues, thereby assisting the model in effectively utilizing contextual information \cite{liu2023pre}. Relative studies \cite{schick2021exploiting, zhou2022prompt} have demonstrated that prompt learning leads to improved performance of pre-trained language models in few-shot and zero-shot scenarios.
For example, the CLIP \cite{radford2021learning} vision-language model leverages textual prompts that explain the concepts present in the image and achieves generalizable cross-modal matching. Similarly, in the speech-language domain, context from text prompts is used to improve speech emotion recognition \cite{jeong2023multimodal}. Given previous works \cite{schick2021exploiting, zhou2022prompt} on NLP that have shown how prompts can aid in improving the model's fit to pre-trained models and better utilize prior knowledge, further research on prompt learning should be a crucial consideration in advancing the efficacy of models within the audio-visual field.

\section{Generalizable Audio-Visual Segmentation}
\label{method}

In practical applications, challenging generalization-related issues such as zero-shot and few-shot segmentation on unseen classes and different datasets can deteriorate the performance of pre-trained audio-visual models. In this section, we introduce how our model, GAVS as shown in Figure \ref{fig:model}, deals with the above issues by focusing on the following two aspects: (i) constructing audio prompts to guide audio source decoding, (ii) correctly projecting audio prompts into the visual space and generating the corresponding mask.

\subsection{Multimodal Representation}
Based on the previous works \cite{zhou2022audio, gao2023avsegformer} for video and audio processing, we sample the video at intervals of 1 second to obtain frames $x_{frames} \in \mathbb{R}^{T\times3\times H\times W}$, where $T$ represents the number of frames as well as the video duration in seconds. The above operations transform the video segmentation task into image segmentation. The visual foundation model SAM extracts image features from a ViT \cite{dosovitskiy2020image} model containing 12 transformer layers. We further tune the visual encoder with bottleneck adapters \cite{houlsby2019parameter} and obtain the visual feature $F_V \in \mathbb{R}^{d_V \times H \times W}$.

We extract audio features using the VGGish \cite{hershey2017cnn} method. The VGGish model is specifically designed for audio feature extraction and is capable of capturing both temporal and spectral information. Firstly, we preprocess the audio into a mono-waveform with a sampling rate of 16kHz. Then, we use the Fourier transform to obtain the mel spectrum, which is subsequently fed into the VGGish model to extract audio feature $F_{As} \in \mathbb{R}^{T \times d_{m}}$ and $d_{m}$ is 128 in default. Finally, for each video clip, the $i_{th}$ frame corresponds to the audio feature $F_{A} = F_{As}[i]$.

\subsection{Semantic-aware Audio Prompting}
SAP prompts the visual foundation model to retrieve sounding objects from the visual space by leveraging the prior knowledge and consists of audio input, visual cues and learnable adaptive noise, as shown in the left part of Figure \ref{fig:model}. The global average pooling $GAP(\cdot)$ is designed to incorporate visual cues into the audio input, thereby introducing visual context to enhance the audio-visual correlation during the audio source decoding.

As shown in Equation \ref{eq:f_cues}, we first obtain the comprehensive visual feature $F_{VG} \in \mathbb{R}^{d_V}$ by performing $GAP$ on the visual feature $F_V$, then we feed $F_{VG}$ into an MLP module to achieve consistent dimension with the audio feature $F_A$, resulting in the visual cues $F_C \in \mathbb{R}^{d_{m}}$:

\begin{equation}
        \label{eq:f_cues}
        F_{C} = MLP(GAP(F_{V})),
\end{equation}
\noindent the reason for unifying the dimensions of visual and audio features is to enable contrastive learning, which extracts cross-modal representations with semantic consistency and thus enhances cross-modal generalization. Besides, incorporating visual cues as scene contextual information for audio input can provide semantic context from the visual modality during cross-modal audio-visual interactions. 

In addition to visual cues, we introduce a learnable adaptive noise $F_N \in \mathbb{R}^{d_N}$ as part of the audio prompt. Instead of explicitly providing semantic information, the adaptive noise prompt implicitly aligns current modality features with the data distribution of the visual foundation model during the tuning process for specific downstream tasks. Moreover, embedding adaptive noise into audio input provides more diverse representations of audio prompts in the feature space, enhancing the model's generalization and noise tolerance during the inference.

Through the aforementioned operations, we simply concatenate the prompt components and audio input to obtain the final audio prompt $F_{A^\prime} \in \mathbb{R}^{2d_m+d_N}$, which we also refer to as SAP:
\begin{equation}
        \label{eq:f_audio_prompt}
        F_{A^\prime} = [F_C;F_N;F_A].
\end{equation}

Finally, we feed the visual input and projected prompt\footnote{SAM provides 6 token slots including 1 IoU token, 4 query tokens and 1 prompt token; we use the first query token for mask generating.} $F_P \in \mathbb{R}^{6 \times d_V}$  into the Audio Source Decoder for sounding object segmentation. 

\subsection{Audio Source Decoder}
In previous approaches, the decoder generates pixel-level masks based on fused features. We argue that decoding in the visual space with the help of audio prompts can enhance the generalization ability of AVS models. Specifically, we tune the mask decoder of SAM. However, to maintain the prior knowledge of the visual foundation model, instead of tuning the whole decoder or modifying the cross-modal attention modules in the middle of Figure \ref{fig:model} that already contain prior interactive knowledge, we propose ColA method to efficiently construct the audio-visual correlation by tuning the core context engaging in different cross-modal attention $CMA(\cdot)$ modules:
\begin{equation}
\label{eq:context} 
    F_{context} = F_P + CMA(F_P, F_{V}^T),
\end{equation}
\begin{equation}
\label{eq:cola}     
    F_{P^\prime} = ColA(MLP(F_{context})) + MLP(F_{context}),
\end{equation}

\noindent where $ColA(\cdot)$ is a bottleneck adapter, and $F_{context} \in \mathbb{R}^{6 \times d_V}$is the addition of $F_P$ and output of AV cross-modal attention (the former one in Figure \ref{fig:model}). $F_{P^\prime}$ is the updated prompt feature that is ready to be fed into VA cross-modal attention (the latter one in Figure \ref{fig:model}) with $F_{context}$, serving as the key $K \in \mathbb{R}^{6 \times d_V}$. Then we get the updated visual feature $F_{V^\prime} \in \mathbb{R}^{H\times W \times d_V}$:
\begin{equation}
    K = F_{context} + F_{P^\prime},
\end{equation}
\begin{equation}
    F_{V^\prime} = F_V^T + CMA(F_V^T, K).
\end{equation}

By employing the above approach, we only need to tune the core context features to establish the outstanding audio-visual correlation. We later validate the effectiveness of ColA in the ablation study by comparing it with tuning the cross-modal attention modules.

After traversing through all transformer layers, we use the final visual output as the mask embedding $F_M \in \mathbb{R}^{d_V \times H \times W}$. Then, we upscale the mask embedding by a transposed convolutional module and we get the upscaled embedding $F_{up} \in \mathbb{R}^{\frac{d_v}{8} \times 4H \times 4W}$. 

Next, we pass the object query through the MLP module and finally, the mask $M_{pred} \in \mathbb{R}^{4H\times 4W}$ is calculated based on the query part of $F_{P}$ and the upscaled embedding:
\begin{equation}
    \label{eq:mask_final}
    M_{pred} = F_{up} \times MLP(F_{P}[1]).
\end{equation}

The above flows accomplish AVS based on audio prompts. Then we can directly add the mask embedding from the two-way transformer to the image embedding $F_V$ for further training.

\begin{table*}[!htb]
    \centering
    \footnotesize
    \scalebox{0.97}{
    \begin{tabular}{lcccccccc}
        \hline
        ~ & ~ & ~ & \multicolumn{2}{c}{V1S} & \multicolumn{2}{c}{V1M} & \multicolumn{2}{c}{V2}  \\ 
        Method & Audio-backbone & Visual-backbone & mIoU(\%) & F-score & mIoU(\%) & F-score & mIoU(\%) & F-score \\ 
    
        \hline
        AVSBench (ECCV'2022) & VGGish & PVT-v2 & 78.70 & 0.879 & 54.00 & 0.645 &  62.45 & 0.756 \\ 
        AVSegFormer (AAAI'2024) & VGGish & PVT-v2 & \textbf{82.06} & 0.899 & 58.36 & 0.693 &  64.34 & 0.759 \\ 
        AVSC (ACMMM'2023) & VGGish & PVT-v2 & 81.29 & 0.886 & 59.50 & 0.657 & - & - \\
        
        AuTR (ArXiv'2023) & VGGish & Swin-base & 80.40 & 0.891 & 56.20 & 0.672 & - & - \\

        AV-SAM (ArXiv'2023) &  ResNet18 & ViT-Base & 40.47 & 0.566 & - & - & - & -\\

        Audio-SAM$^\dagger$ (ours) & VGGish & ViT-Base & 56.33 & 0.727 & 33.68 & 0.459 & 57.41 & 0.684  \\

        SAM-Fusion$^\ddagger$ (ours) & VGGish & ViT-Base & 71.92 & 0.775 & 50.61 & 0.637 & 60.19 & 0.724 \\

        \textbf{GAVS (ours)} & VGGish & ViT-Base & 80.06 & \textbf{0.902} & \textbf{63.70} & \textbf{0.774}   & \textbf{67.70} & \textbf{0.788} \\ 
        
        \hline
    \end{tabular}}
    
    \caption{Performance on AVS-Benchmarks. Although GAVS shows deficiencies on the V1S dataset, it exhibits comparable improvements in the V1M and V2 datasets to other models that trained with the encoder-fusion-decoder paradigm. $\dagger$: We only replace the sparse prompt in SAM with audio inputs, to conduct a comparative experiment with AV-SAM. $\ddagger$: Set up similar to GAVS, but fuse the audio and visual modalities without prompting before the Audio Source Decoder.}
    \label{tab:full-avs}
\end{table*}

\begin{figure*}[!htb]
     \centering
     \includegraphics[width=0.76\textwidth]{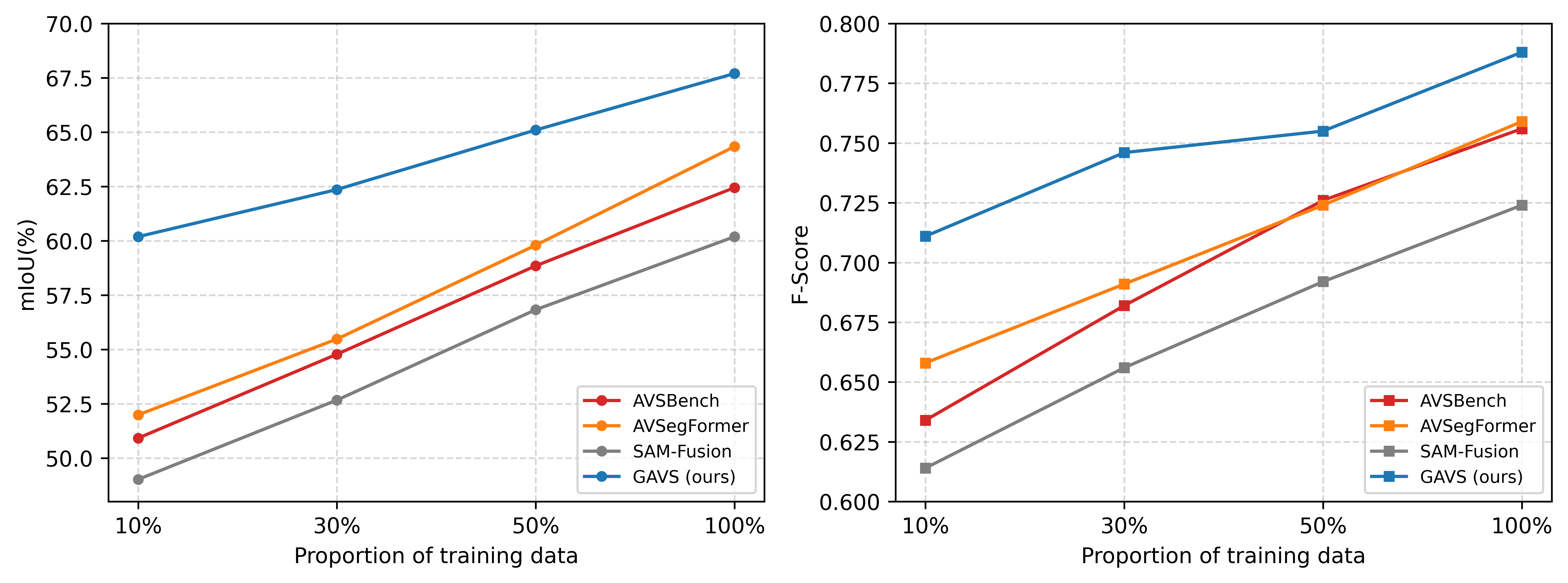}
     \vspace{-0.5em}
     \caption{Visualization of performance improvements of AVS models on the AVS-V2 dataset in relation to the amount of data used for training. We compare models with subsets consisting of 10\%, 30\%, and 50\% of the full dataset. Our results show that our method achieves better performance with only 10\% of the training data compared to other models trained with 30\%. Moreover, our model outperforms other models trained on the full dataset when trained with only half of the data.}
     \label{fig:v2_part}
     \vspace{-0.75em}
\end{figure*}

\subsection{Learning Objectives}
\paragraph{Segmentation Loss.}We use the binary cross-entropy $BCE(\cdot)$ loss to measure the difference between the model's predicted mask and the ground truth label during the model training process:
\begin{equation}
\label{eq:loss_seg}
\mathcal{L}_{seg} = \text{BCE}(M_{pred}, M_{gt}).
\end{equation}

\paragraph{Semantic Loss.} We adopt a simple triplet loss to optimize contrastive learning for achieving semantic alignment, and use cosine similarity as the metric for feature similarity measurement. For each video, we select the average visual feature $v_i$ and the average audio feature $a_i$ as the positive pair, and select the average visual feature $v_i$ and the average audio feature $a_j$ of the other video as the negative pair:

\begin{equation}
\label{eq:avg_av}
v_i = \bar{F_V} = \frac{\sum_{}^{T}F_V}{T}; a_i = \bar{F_A} = \frac{\sum_{}^{T}F_A}{T},
\end{equation}
\begin{equation}
\label{eq:loss_semantic}
\mathcal{L}_{sem} = \frac{1}{N}\sum_{i=1}^{N}\left[m - \text{sim}(v_i,a_i) + \max_{j=1}^{N}[ \text{sim}(v_i,a_j)]\right]_+,
\end{equation}
in which $N$ represents the mini-batch size, and $m$ represents the margin to control the distance between positive pairs and negative pairs.

\paragraph{Total Loss.} 
The final loss function is a linear combination of the aforementioned loss functions:
\begin{equation}
\label{eq:tot_loss}
\mathcal{L} = \mathcal{L}_{seg} + \lambda \mathcal{L}_{sem},
\end{equation}
where $\lambda$ is the weight of semantic loss.

\section{Experiments}
\label{experiments}
To evaluate the grounding performance of our model, we conduct tests on AVS-Benchmarks and use mean intersection over union (mIoU) and F-score as the performance metrics, following previous works \cite{zhou2022audio, gao2023avsegformer}. Additionally, to assess the generalization ability, we split zero-shot and few-shot testing subsets\footnote{Refer to the project page for detailed split settings.} based on AVS-Benchmarks and VGG-SS datasets.


\subsection{Grounding Segmentation on AVS-Benchmarks}

AVS-Benchmarks \cite{zhou2022audio} is a dataset specifically designed for AVS tasks. Refer to Table \ref{tab:full-avs}, our model achieves the best performance in multi-source setting (V1M and V2) and gets comparable performance in single-source setting (V1S). Compared with AV-SAM, where both models utilize prompts, our implemented straightforward Audio-SAM freezes all parameters except for the audio input, which is passed through an additional MLP module for updating. This results in a performance improvement of 15\% compared to AV-SAM, demonstrating the effectiveness of the encoder-prompt-decoder paradigm, which directly prompted the visual foundation model.

Besides, we further compare the performance of various open-source models at different data volumes to demonstrate our superiority in data utilization, as the AVS task is cost-intensive. As shown in Figure \ref{fig:v2_part}, with only 50\% of the data, we can achieve the best performance equivalent to using 100\% of the data by other models.

\begin{table*}[!htbp]
\centering
\footnotesize
\scalebox{1}{
\begin{tabular}{lccccccccccc}
\hline
& \multicolumn{2}{c}{0-shot} & \multicolumn{2}{c}{1-shot} & \multicolumn{2}{c}{3-shot} & \multicolumn{2}{c}{5-shot} \\ 

Method & mIoU(\%)     & F-score  & mIoU(\%)     & F-score   & mIoU(\%)     & F-score     & mIoU(\%)     & F-score   \\ 

\hline
AVSBench (ECCV'2022) & 53.00 & 0.707 & 56.11 & 0.754 & 63.22 & 0.767 & 63.87 & 0.783 \\

AVSegFormer (ArXiv'2023) & 54.26 & 0.715 & 58.30 & 0.764 & 64.19 & 0.774 & 65.17 & 0.785 \\ 

SAM-Fusion (ours) & 46.25 & 0.630 & 50.39 & 0.671 & 57.05 & 0.719 & 60.82 & 0.741 \\ 


\textbf{GAVS (ours)} & \textbf{54.71} & \textbf{0.722} & \textbf{62.89} & \textbf{0.768}  & \textbf{66.28} & \textbf{0.774} & \textbf{67.75} & \textbf{0.795} \\ 
\hline
\end{tabular}
}
\vspace{-0.5em}
\caption{Performance on AVS-V3 for testing the generalization ability on unseen object classes. Our model GAVS, which is trained with the encoder-prompt-decoder paradigm achieves a significant performance improvement compared to other models following the conventional encoder-fusion-decoder paradigm. 
}
\vspace{-0.5em}
\label{tab:avs-v3}
\end{table*}
\subsection{Unseen Classes on AVS-V3}
We design AVS-V3 to assess the generalization ability of AVS models on unseen classes. It consists of four settings: 0-shot, 1-shot, 3-shot, and 5-shot. In the zero-shot setting, the classes of objects in the test set do not appear during the training or validation. In the other settings, we select $N$=[1, 3, 5] data samples for each class and include them in the training process to enable few-shot learning.

As shown in Table \ref{tab:avs-v3}, our model achieves the highest 0-shot performance, exhibiting superior generalization when encountering unseen object classes. Meanwhile, we can observe that after 3-shot learning, our model surpasses other models' performance trained with 5-shot, indicating that our model possesses better few-shot learning ability.


\begin{table}[!tb]
\centering
\footnotesize
\scalebox{1}{
\begin{tabular}{lccc}
\hline
Method & Train & cIoU(\%)   & AUC \\
\hline
HardWay (CVPR'2021) & in-set & 34.4 & 0.382 \\
EZ-VSL (ECCV'2022) & in-set & 38.85  & 0.395 \\
SLAVC (NeurIPS'2022) & in-set & 39.80  & -  \\ 
MarginNCE (ICASSP'2023) & in-set & 39.78 & 0.400  \\
AVIN-RN (ACMMM'2023) & in-set & \textbf{44.90} & \textbf{0.436} \\

\hline
AVSBench (ECCV'2022) & zero-shot & 36.86 & 0.370 \\
AVSegFormer (ArXiv'2023) & zero-shot & 38.86 & 0.390 \\

SAM-Fusion (ours)  & zero-shot &  30.17  &  0.302       \\
\textbf{GAVS (ours)} & zero-shot & \textbf{41.07} & \textbf{0.411}     \\ 
\hline
\end{tabular}
}
\caption{The results of VGG-SS for comparing the performance of zero-shot AVS models with traditional self-supervised in-set AVL models. Our model outperforms other AVS models in cross-dataset settings.}
\label{lab:V2-VGG_SS}
\vspace{-0.6em}
\end{table}

\subsection{Cross-datasets on VGG-SS}
\paragraph{VGG-SS.} VGG-SS \cite{chen2021localizing} is a dataset designed for the AVL task performance test. Each image has a corresponding audio source and a bounding box label. VGG-SS contains 5,158 images covering 220 categories, and all the data is only used for testing purposes. 

In this experiment, we test models' cross-dataset generalization on the VGG-SS test set. Previous works  such as HardWay \cite{chen2021localizing}, EZ-VSL \cite{mo2022localizing}, SLAVC \cite{mo2022closer}, MarginNCE \cite{park2023marginnce} and AVIN-RN \cite{liu2023induction} trained models on VGG-Sound 144k, we label them as ``trained on in-set" because VGG-SS is extracted from VGG-Sound. In contrast, we train typical AVS models on AVS-V2 and can be labelled as ``trained with zero-shot" for cross-dataset testing. As shown in Table \ref{lab:V2-VGG_SS}, models such as AVSBench and AVSegFormer perform well on AVS-Benchmarks but fail to perform as well in VGG-SS. Our model has better cross-dataset generalization ability and surpasses other zero-shot models, although there is still some gap compared to the best in-set model.

\paragraph{VGG-SS-Sub.} Due to VGG-SS only containing the test set, we split it and obtain VGG-SS-Sub to assess the few-shot cross-dataset generalization ability of fusion-based and prompt-based AVS models transfer from AVS to AVL task. Same with the AVS-V3, it is set up with zero-shot and few-shot (1, 3, 5) settings. Note that the zero-shot performance of this subset cannot be compared with the VGG-SS full set as the test set is different.

From Table \ref{tab:vgg-ss-sub}, we can observe that our model achieves better zero-shot and few-shot performance, suggesting that with SAP and ColA, our model can better fit the data distribution across different datasets.

\begin{table*}[!tb]
\centering
\footnotesize
\scalebox{1}{
\begin{tabular}{lcccccccc}
\hline
& \multicolumn{2}{c}{0-shot} & \multicolumn{2}{c}{1-shot} & \multicolumn{2}{c}{3-shot} & \multicolumn{2}{c}{5-shot} \\ 

Method & cIoU(\%) & AUC  & cIoU(\%) & AUC & cIoU(\%) &  AUC  & cIoU(\%) &  AUC \\ 

\hline
AVSBench (ECCV'2022) & 37.28 & 0.374 & 53.33 & 0.534 & 56.78 & 0.569 & 57.38 & 0.574 \\

AVSegFormer (ArXiv'2023) & 37.99 & 0.380 & 53.41 & 0.534 & 56.84 & 0.569 & 57.65 & 0.577 \\ 

SAM-Fusion (ours) & 31.22 & 0.315 & 40.39 & 0.407 & 45.25 & 0.453 & 48.67 & 0.487 \\ 


\textbf{GAVS (ours)} & \textbf{38.62} & \textbf{0.387} & \textbf{53.70} & \textbf{0.537} & \textbf{57.41} & \textbf{0.574} & \textbf{60.14} & \textbf{0.602} \\ 
\hline
\end{tabular}
}
\vspace{-0.5em}
\caption{Performance on VGG-SS-Sub for testing the generalization ability across different datasets. Our model is trained following the encoder-prompt-decoder paradigm and achieves the best zero-shot and few-shot performance.}
\label{tab:vgg-ss-sub}
\end{table*}

\begin{figure*}[!tb]
     \centering 
     \includegraphics[width=0.85\textwidth]{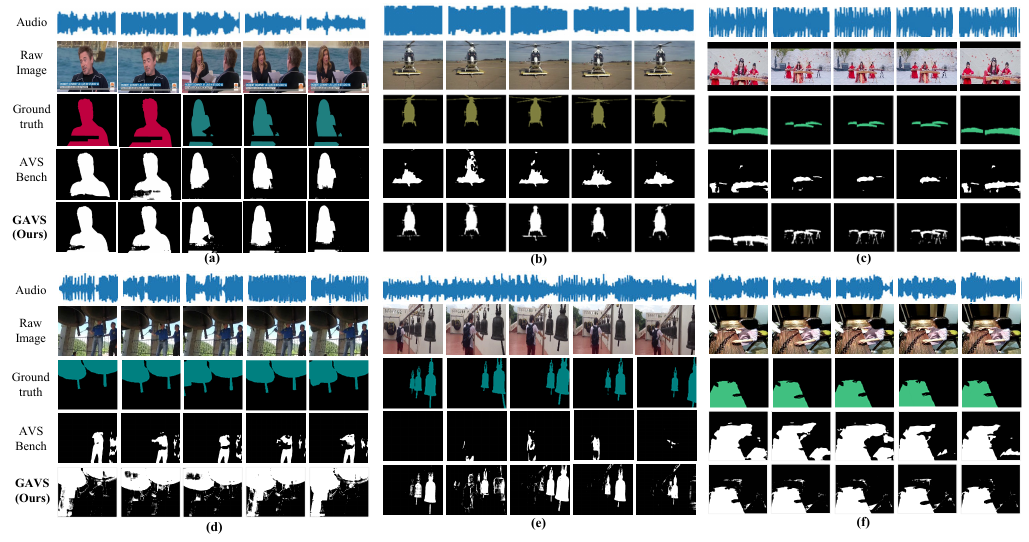} 
     \caption{Visualization of segmented masks on the AVS-V2 (a$\sim$c) and AVS-V3 (d$\sim$f) zero-shot test set for unseen classes. The results reveal that our method effectively identifies objects although their semantic classes were not present in the training set. This observation suggests that our model possesses superior zero-shot generalization capabilities compared to AVSBench, which follows the encoder-fusion-decoder paradigm.}
     \label{fig:qualitative}
     \vspace{-0.5em}
\end{figure*}

\subsection{Ablation Study}
\vspace{-0.25em}
As shown in Table \ref{tab:ablation}, we evaluate the capabilities of our model on the AVS-V2 dataset. Initially, We investigate the effects of different tuning strategies in the Audio Source Decoder to establish the correlation between audio and visual modalities. We conduct separate tests for a) freezing the decoder parameters, b) fine-tuning the entire decoder, and c) tuning the AV cross-modal attention and d) VA cross-modal attention using adapters, comparing them with our proposed e) ColA, and the combined strategy f) ColA \& tuning AV and VA cross-modal attention with adapters. Experimental results demonstrate that adapter-based tuning outperforms freezing and fine-tuning. Additionally, our proposed ColA achieves better results compared to modifying cross-modal attention components that include pre-trained knowledge, indicating that ColA can better construct the audio-visual correlation by leveraging the prior knowledge of the visual foundation model. 

Furthermore, we explore the effectiveness of different AVS paradigms with the same visual foundation model, and the results of g) and h) indicate that using audio as cues to prompt the visual foundation model can outperform fusing audio and visual modalities directly. Building upon the encoder-prompt-decoder paradigm, we further incorporate i) visual backbone adapters and j) SAP, resulting in improvements in segmentation performance.

\begin{table}[!tb]
\centering
\footnotesize
\scalebox{1}{
\begin{tabular}{clccc}
\hline
& Method  & mIoU(\%)   & F-score \\
\hline
a) & freeze & 57.41 & 0.684 \\
b) & fine-tune  & 62.08  & 0.714  \\
c) & AV-adapter & 62.11 & 0.720 \\
d) & VA-adapter & 61.84 & 0.715 \\
e) & ColA   & 63.35  & 0.727   \\
f) & ColA + AV + VA & 62.09 &  0.721\\
\hline
g) & AV-fusion & 59.69 & 0.701 \\
h) & audio-prompt  & 63.35 & 0.727 \\
\hline
i) & +visual-adapter & 65.92 & 0.759  \\
j) & +SAP & 67.70 & 0.788 \\

\hline
\end{tabular}
}
\caption{Ablation study. We conduct ablation analyses on the AVS-V2 dataset and the results show that adding visual-adapters and SAP both contribute to better performance gains compared to using only audio prompts. We also demonstrate the superiority of ColA in building audio-visual correlation compared to other tuning methods.}
\label{tab:ablation}
\vspace{-1.3em}
\end{table}

\subsection{Qualitative Analysis}
As shown in Figure \ref{fig:qualitative}, (a)(b)(c) represent the visualization results on AVS-V2, while (d)(e)(f) represent the visualization results on AVS-V3 zero-shot test set. On AVS-V2, our model produces higher-quality visualization results, indicating that our model has better grounding AVS performance. It is worth noting that in AVS-V3, the object classes in the test set were not included during the training phase. Our model, trained using the encoder-prompt-decoder paradigm, exhibits improved object recognition ability for unseen classes when prompted with audio input. In contrast, AVSBench erroneously segments or incorrectly recognizes extra objects.

\section{Conclusion and Future Work}
\vspace{-0.5em}
\label{conclusion}
The development of large pre-trained models has greatly enhanced the generalization performance of traditional CV tasks, but little attention is given to the generalization of cross-modal AVS in zero-shot and few-shot scenarios. In this work, we introduce GAVS, the model following the encoder-prompt-decoder paradigm to address the increasing demand for precise localization with limited annotated data and varying data distribution in real-world scenarios. Our method achieves generalizable cross-modal segmentation, benefiting from using SAP to help the visual foundation model focus on the sounding objects and using ColA for efficient audio-visual correlation construction. Our method is only one solution and provides a reference for exploring generalizable AVS, future work can investigate more flexible methods for generalizable audio-visual correlation learning based on large pre-trained models, as well as how to effectively handle the interaction between audio and visual features to further promote the model's generalization.

\vspace{-0.5em}
\section{Acknowledgments}
This research was supported by the National Natural Science Foundation of China (NO.62106272), the Young Elite Scientists Sponsorship Program by CAST (2021QNRC001), the National Natural Science Foundation of China under Grant U20A20222, the National Science Foundation for Distinguished Young Scholars under Grant 62225605 and the Public Computing Cloud, Renmin University of China. 

\bibliography{aaai24}

\begin{thebibliography}{36}
\providecommand{\natexlab}[1]{#1}

\bibitem[{Arandjelovic and Zisserman(2018)}]{arandjelovic2018objects}
Arandjelovic, R.; and Zisserman, A. 2018.
\newblock Objects that sound.
\newblock In \emph{Proceedings of the European conference on computer vision (ECCV)}, 435--451.

\bibitem[{Baltru{\v{s}}aitis, Ahuja, and Morency(2018)}]{baltruvsaitis2018multimodal}
Baltru{\v{s}}aitis, T.; Ahuja, C.; and Morency, L.-P. 2018.
\newblock Multimodal machine learning: A survey and taxonomy.
\newblock \emph{IEEE transactions on pattern analysis and machine intelligence}, 41(2): 423--443.

\bibitem[{Brown et~al.(2020)Brown, Mann, Ryder, Subbiah, Kaplan, Dhariwal, Neelakantan, Shyam, Sastry, Askell et~al.}]{brown2020language}
Brown, T.; Mann, B.; Ryder, N.; Subbiah, M.; Kaplan, J.~D.; Dhariwal, P.; Neelakantan, A.; Shyam, P.; Sastry, G.; Askell, A.; et~al. 2020.
\newblock Language models are few-shot learners.
\newblock \emph{Advances in neural information processing systems}, 33: 1877--1901.

\bibitem[{Chen et~al.(2021)Chen, Xie, Afouras, Nagrani, Vedaldi, and Zisserman}]{chen2021localizing}
Chen, H.; Xie, W.; Afouras, T.; Nagrani, A.; Vedaldi, A.; and Zisserman, A. 2021.
\newblock Localizing visual sounds the hard way.
\newblock In \emph{Proceedings of the IEEE/CVF Conference on Computer Vision and Pattern Recognition}, 16867--16876.

\bibitem[{Chen et~al.(2020)Chen, Xie, Vedaldi, and Zisserman}]{chen2020vggsound}
Chen, H.; Xie, W.; Vedaldi, A.; and Zisserman, A. 2020.
\newblock VGGSound: A Large-scale Audio-Visual Dataset.
\newblock arXiv:2004.14368.

\bibitem[{Dosovitskiy et~al.(2020)Dosovitskiy, Beyer, Kolesnikov, Weissenborn, Zhai, Unterthiner, Dehghani, Minderer, Heigold, Gelly et~al.}]{dosovitskiy2020image}
Dosovitskiy, A.; Beyer, L.; Kolesnikov, A.; Weissenborn, D.; Zhai, X.; Unterthiner, T.; Dehghani, M.; Minderer, M.; Heigold, G.; Gelly, S.; et~al. 2020.
\newblock An image is worth 16x16 words: Transformers for image recognition at scale.
\newblock \emph{arXiv preprint arXiv:2010.11929}.

\bibitem[{Gao et~al.(2023)Gao, Chen, Chen, Wang, and Lu}]{gao2023avsegformer}
Gao, S.; Chen, Z.; Chen, G.; Wang, W.; and Lu, T. 2023.
\newblock AVSegFormer: Audio-Visual Segmentation with Transformer.
\newblock \emph{arXiv preprint arXiv:2307.01146}.

\bibitem[{He et~al.(2016)He, Zhang, Ren, and Sun}]{he2016deep}
He, K.; Zhang, X.; Ren, S.; and Sun, J. 2016.
\newblock Deep residual learning for image recognition.
\newblock In \emph{Proceedings of the IEEE conference on computer vision and pattern recognition}, 770--778.

\bibitem[{Hershey et~al.(2017)Hershey, Chaudhuri, Ellis, Gemmeke, Jansen, Moore, Plakal, Platt, Saurous, Seybold et~al.}]{hershey2017cnn}
Hershey, S.; Chaudhuri, S.; Ellis, D.~P.; Gemmeke, J.~F.; Jansen, A.; Moore, R.~C.; Plakal, M.; Platt, D.; Saurous, R.~A.; Seybold, B.; et~al. 2017.
\newblock CNN architectures for large-scale audio classification.
\newblock In \emph{2017 ieee international conference on acoustics, speech and signal processing (icassp)}, 131--135. IEEE.

\bibitem[{Houlsby et~al.(2019)Houlsby, Giurgiu, Jastrzebski, Morrone, De~Laroussilhe, Gesmundo, Attariyan, and Gelly}]{houlsby2019parameter}
Houlsby, N.; Giurgiu, A.; Jastrzebski, S.; Morrone, B.; De~Laroussilhe, Q.; Gesmundo, A.; Attariyan, M.; and Gelly, S. 2019.
\newblock Parameter-efficient transfer learning for NLP.
\newblock In \emph{International Conference on Machine Learning}, 2790--2799. PMLR.

\bibitem[{Hu et~al.(2021)Hu, Wei, Qian, Lin, Song, and Wen}]{hu2021class}
Hu, D.; Wei, Y.; Qian, R.; Lin, W.; Song, R.; and Wen, J.-R. 2021.
\newblock Class-aware sounding objects localization via audiovisual correspondence.
\newblock \emph{IEEE Transactions on Pattern Analysis and Machine Intelligence}, 44(12): 9844--9859.

\bibitem[{Jeong, Kim, and Kang(2023)}]{jeong2023multimodal}
Jeong, E.; Kim, G.; and Kang, S. 2023.
\newblock Multimodal Prompt Learning in Emotion Recognition Using Context and Audio Information.
\newblock \emph{Mathematics}, 11(13): 2908.

\bibitem[{Jia and Zhang(2022)}]{jia2022prompt}
Jia, C.; and Zhang, Y. 2022.
\newblock Prompt-based Distribution Alignment for Domain Generalization in Text Classification.
\newblock In \emph{Proceedings of the 2022 Conference on Empirical Methods in Natural Language Processing}, 10147--10157.

\bibitem[{Kirillov et~al.(2023)Kirillov, Mintun, Ravi, Mao, Rolland, Gustafson, Xiao, Whitehead, Berg, Lo et~al.}]{kirillov2023segment}
Kirillov, A.; Mintun, E.; Ravi, N.; Mao, H.; Rolland, C.; Gustafson, L.; Xiao, T.; Whitehead, S.; Berg, A.~C.; Lo, W.-Y.; et~al. 2023.
\newblock Segment anything.
\newblock \emph{arXiv preprint arXiv:2304.02643}.

\bibitem[{Lester, Al-Rfou, and Constant(2021)}]{lester2021power}
Lester, B.; Al-Rfou, R.; and Constant, N. 2021.
\newblock The Power of Scale for Parameter-Efficient Prompt Tuning.
\newblock In \emph{Proceedings of the 2021 Conference on Empirical Methods in Natural Language Processing}, 3045--3059.

\bibitem[{Li et~al.(2022)Li, Zhuang, Fan, and Wang}]{li2022learning}
Li, A.; Zhuang, L.; Fan, S.; and Wang, S. 2022.
\newblock Learning common and specific visual prompts for domain generalization.
\newblock In \emph{Proceedings of the Asian Conference on Computer Vision}, 4260--4275.

\bibitem[{Li and Liang(2021)}]{li2021prefix}
Li, X.~L.; and Liang, P. 2021.
\newblock Prefix-Tuning: Optimizing Continuous Prompts for Generation.
\newblock In \emph{Proceedings of the 59th Annual Meeting of the Association for Computational Linguistics and the 11th International Joint Conference on Natural Language Processing (Volume 1: Long Papers)}, 4582--4597.

\bibitem[{Ling et~al.(2023)Ling, Li, Gan, Zhang, Chi, and Wang}]{ling2023hear}
Ling, Y.; Li, Y.; Gan, Z.; Zhang, J.; Chi, M.; and Wang, Y. 2023.
\newblock Hear to Segment: Unmixing the Audio to Guide the Semantic Segmentation.
\newblock \emph{arXiv preprint arXiv:2305.07223}.

\bibitem[{Liu et~al.(2023{\natexlab{a}})Liu, Li, Qi, Zhang, Li, Wang, and Yu}]{liu2023audiovisual}
Liu, C.; Li, P.; Qi, X.; Zhang, H.; Li, L.; Wang, D.; and Yu, X. 2023{\natexlab{a}}.
\newblock Audio-Visual Segmentation by Exploring Cross-Modal Mutual Semantics.
\newblock arXiv:2307.16620.

\bibitem[{Liu et~al.(2023{\natexlab{b}})Liu, Ju, Ma, Wang, Wang, and Zhang}]{liu2023audioaware}
Liu, J.; Ju, C.; Ma, C.; Wang, Y.; Wang, Y.; and Zhang, Y. 2023{\natexlab{b}}.
\newblock Audio-aware Query-enhanced Transformer for Audio-Visual Segmentation.
\newblock \emph{arXiv preprint arXiv:2307.13236}.

\bibitem[{Liu et~al.(2023{\natexlab{c}})Liu, Yuan, Fu, Jiang, Hayashi, and Neubig}]{liu2023pre}
Liu, P.; Yuan, W.; Fu, J.; Jiang, Z.; Hayashi, H.; and Neubig, G. 2023{\natexlab{c}}.
\newblock Pre-train, prompt, and predict: A systematic survey of prompting methods in natural language processing.
\newblock \emph{ACM Computing Surveys}, 55(9): 1--35.

\bibitem[{Liu et~al.(2023{\natexlab{d}})Liu, Zhang, Huang, Zha, You, and Zhang}]{liu2023induction}
Liu, T.; Zhang, P.; Huang, W.; Zha, Y.; You, T.; and Zhang, Y. 2023{\natexlab{d}}.
\newblock Induction Network: Audio-Visual Modality Gap-Bridging for Self-Supervised Sound Source Localization.
\newblock \emph{arXiv preprint arXiv:2308.04767}.

\bibitem[{Mo and Morgado(2022{\natexlab{a}})}]{mo2022closer}
Mo, S.; and Morgado, P. 2022{\natexlab{a}}.
\newblock A closer look at weakly-supervised audio-visual source localization.
\newblock \emph{Advances in Neural Information Processing Systems}, 35: 37524--37536.

\bibitem[{Mo and Morgado(2022{\natexlab{b}})}]{mo2022localizing}
Mo, S.; and Morgado, P. 2022{\natexlab{b}}.
\newblock Localizing visual sounds the easy way.
\newblock In \emph{European Conference on Computer Vision}, 218--234. Springer.

\bibitem[{Mo and Tian(2023)}]{mo2023av}
Mo, S.; and Tian, Y. 2023.
\newblock AV-SAM: Segment anything model meets audio-visual localization and segmentation.
\newblock \emph{arXiv preprint arXiv:2305.01836}.

\bibitem[{Park, Senocak, and Chung(2023)}]{park2023marginnce}
Park, S.; Senocak, A.; and Chung, J.~S. 2023.
\newblock Marginnce: Robust sound localization with a negative margin.
\newblock In \emph{ICASSP 2023-2023 IEEE International Conference on Acoustics, Speech and Signal Processing (ICASSP)}, 1--5. IEEE.

\bibitem[{Radford et~al.(2021)Radford, Kim, Hallacy, Ramesh, Goh, Agarwal, Sastry, Askell, Mishkin, Clark et~al.}]{radford2021learning}
Radford, A.; Kim, J.~W.; Hallacy, C.; Ramesh, A.; Goh, G.; Agarwal, S.; Sastry, G.; Askell, A.; Mishkin, P.; Clark, J.; et~al. 2021.
\newblock Learning transferable visual models from natural language supervision.
\newblock In \emph{International conference on machine learning}, 8748--8763. PMLR.

\bibitem[{Schick and Sch{\"u}tze(2021)}]{schick2021exploiting}
Schick, T.; and Sch{\"u}tze, H. 2021.
\newblock Exploiting Cloze-Questions for Few-Shot Text Classification and Natural Language Inference.
\newblock In \emph{Proceedings of the 16th Conference of the European Chapter of the Association for Computational Linguistics: Main Volume}, 255--269.

\bibitem[{Senocak et~al.(2018)Senocak, Oh, Kim, Yang, and Kweon}]{senocak2018learning}
Senocak, A.; Oh, T.-H.; Kim, J.; Yang, M.-H.; and Kweon, I.~S. 2018.
\newblock Learning to localize sound source in visual scenes.
\newblock In \emph{Proceedings of the IEEE Conference on Computer Vision and Pattern Recognition}, 4358--4366.

\bibitem[{Shu et~al.(2022)Shu, Nie, Huang, Yu, Goldstein, Anandkumar, and Xiao}]{shu2022test}
Shu, M.; Nie, W.; Huang, D.-A.; Yu, Z.; Goldstein, T.; Anandkumar, A.; and Xiao, C. 2022.
\newblock Test-time prompt tuning for zero-shot generalization in vision-language models.
\newblock \emph{Advances in Neural Information Processing Systems}, 35: 14274--14289.

\bibitem[{Wei et~al.(2022)Wei, Hu, Tian, and Li}]{wei2022learning}
Wei, Y.; Hu, D.; Tian, Y.; and Li, X. 2022.
\newblock Learning in audio-visual context: A review, analysis, and new perspective.
\newblock \emph{arXiv preprint arXiv:2208.09579}.

\bibitem[{Yang et~al.(2023)Yang, Nachum, Du, Wei, Abbeel, and Schuurmans}]{yang2023foundation}
Yang, S.; Nachum, O.; Du, Y.; Wei, J.; Abbeel, P.; and Schuurmans, D. 2023.
\newblock Foundation models for decision making: Problems, methods, and opportunities.
\newblock \emph{arXiv preprint arXiv:2303.04129}.

\bibitem[{Zang et~al.(2022)Zang, Li, Zhou, Huang, and Loy}]{zang2022unified}
Zang, Y.; Li, W.; Zhou, K.; Huang, C.; and Loy, C.~C. 2022.
\newblock Unified vision and language prompt learning.
\newblock \emph{arXiv preprint arXiv:2210.07225}.

\bibitem[{Zheng et~al.(2022)Zheng, Yue, Wang, and You}]{zheng2022prompt}
Zheng, Z.; Yue, X.; Wang, K.; and You, Y. 2022.
\newblock Prompt vision transformer for domain generalization.
\newblock \emph{arXiv preprint arXiv:2208.08914}.

\bibitem[{Zhou et~al.(2022{\natexlab{a}})Zhou, He, Ma, Berg-Kirkpatrick, and Neubig}]{zhou2022prompt}
Zhou, C.; He, J.; Ma, X.; Berg-Kirkpatrick, T.; and Neubig, G. 2022{\natexlab{a}}.
\newblock Prompt Consistency for Zero-Shot Task Generalization.
\newblock In \emph{Findings of the Association for Computational Linguistics: EMNLP 2022}, 2613--2626.

\bibitem[{Zhou et~al.(2022{\natexlab{b}})Zhou, Wang, Zhang, Sun, Zhang, Birchfield, Guo, Kong, Wang, and Zhong}]{zhou2022audio}
Zhou, J.; Wang, J.; Zhang, J.; Sun, W.; Zhang, J.; Birchfield, S.; Guo, D.; Kong, L.; Wang, M.; and Zhong, Y. 2022{\natexlab{b}}.
\newblock Audio--visual segmentation.
\newblock In \emph{European Conference on Computer Vision}, 386--403. Springer.

\end{thebibliography}
\newpage ~\\~
\newpage
\appendix

\setcounter{table}{0}
\setcounter{figure}{0}

\paragraph{Appendix}

\section*{A Datasets}
\label{datasets}
In this section, we provide a brief introduction to the datasets used in our study.

\subsection{A.1 AVS-Benchmarks}
AVS-Benchmarks \cite{zhou2022audio} is designed for the Audio-Visual Segmentation (AVS) task, and includes three datasets: V1S, V1M, and V2, as shown in Table \ref{tab:avs-stat}.

\begin{table}[!htb]
\centering
\scalebox{0.7}{
\begin{tabular}{lccccc}
    \hline
    ~ & class & videos & train/val/test & frames per video & sound  \\ 
    \hline

    V1S & 23 & 4,932  & 3,452/740/740 & 1/5/5 &  single-source \\
      
    V1M & 23 & 424  & 296/64/64 & 5/5/5 &  multi-source \\
    
    V2 & 70 & 6,000 & 4,750/500/750 & 10/10/10 &  multi-source \\
    \hline
\end{tabular}
}
\caption{Statistics of AVS-Benchmarks. Both V1M and V2 are multi-source AVS while V1S is single-source. V1S is set as a self-supervised task as only the first frame of training set is labelled while all the frame mask should be predicted during the validation and testing.}
\label{tab:avs-stat}
\end{table}

\subsection{A.2 AVS-V3}

To evaluate the model's generalizable segmentation ability for unseen objects, we merge V1M and V2 datasets, resulting in the V3 dataset containing multiple sound sources. Then we extract a set of object categories as the test classes, as shown in Table \ref{tab:v3-stat}. We use the remaining categories as the training data.

\begin{table}[!htb]
\centering
\scalebox{0.58}{
\begin{tabular}{lccccccc}
    \hline
    ~ & Human & Animal & Furniture & Instrument & Transport & SUM & ~ \\ 
    \hline
    
    \multirow{2}{*}{Class} & \multirow{2}{*}{baby} & squirrel, tiger, hen,  & \multirow{2}{*}{bell, clock} & \multirow{2}{*}{guzheng, harp} & helicopter, & \multirow{2}{*}{-} & ~ \\
    ~ & ~ & leopard, donkey  & ~ & ~ & emergency-car & ~ & ~ \\
    Count & 62 & 233 &  125 &  166 &  128 &  714 & ~ \\
    \hline
\end{tabular}
}
\caption{Statistics on AVS-V3 test set. We select 12 object categories from 6 major groups. There are some sounding objects in the original dataset that belong to the same major groups, such as ``man, girl", ``guitar, piano" and ``truck, airplane". We conduct experiments on this dataset to test whether the model can generalize the knowledge learned from the training set to unseen objects in the test set. }
\label{tab:v3-stat}
\end{table}

\subsection{A.3 VGG-SS and VGG-SS-Sub}

VGG-SS dataset \cite{chen2021localizing} comprises 5158 pairs of image-audio, covering 220 categories and is extracted from VGG-Sound dataset. The image annotations are in the format of bounding boxes {Xmin, Ymin, Xmax, Ymax}. As VGG-SS only provides a test set, to conduct few-shot learning, we randomly extract 5 objects per category, while the remaining data is used as the test set for VGG-SS-Sub. In the 0-shot, 1-shot, 3-shot, and 5-shot scenarios, we employ the objects in a corresponding number of shots as the training data. We perform all experiments on VGG-SS and its subsets using our model pre-trained on AVS-V2 to assess the cross-dataset generalization capability of our model. 

\section*{B Visualization}

\begin{figure}[!htbp]
     \includegraphics[width=0.47\textwidth]
     {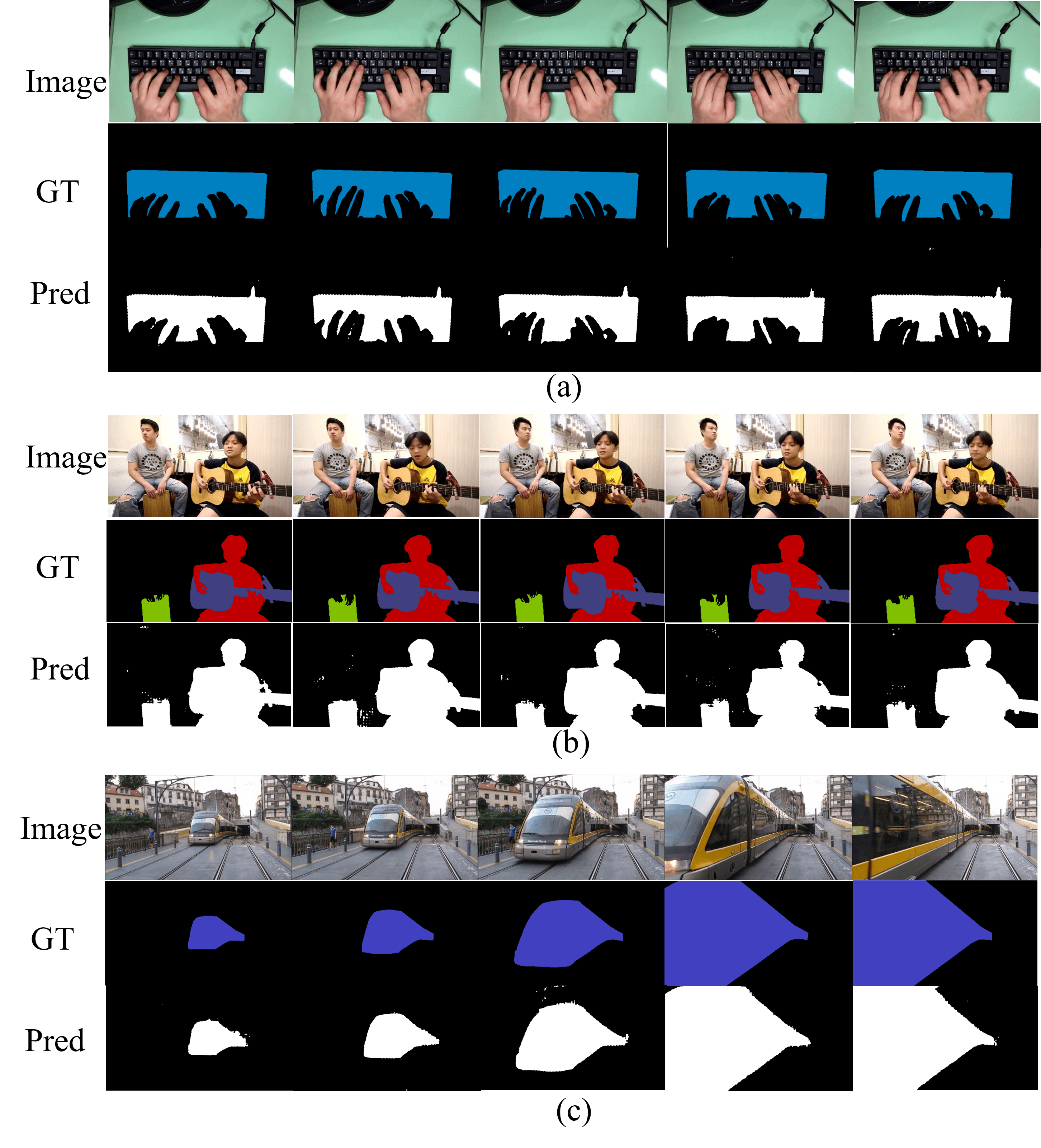}
     \caption{Visualization of segmentation results on the supervised AVS-V2 dataset.}
     \label{fig:3_vis_v2}
\end{figure}

\begin{figure*}[!htb]
     \centering     
     \includegraphics[width=1\textwidth]
     {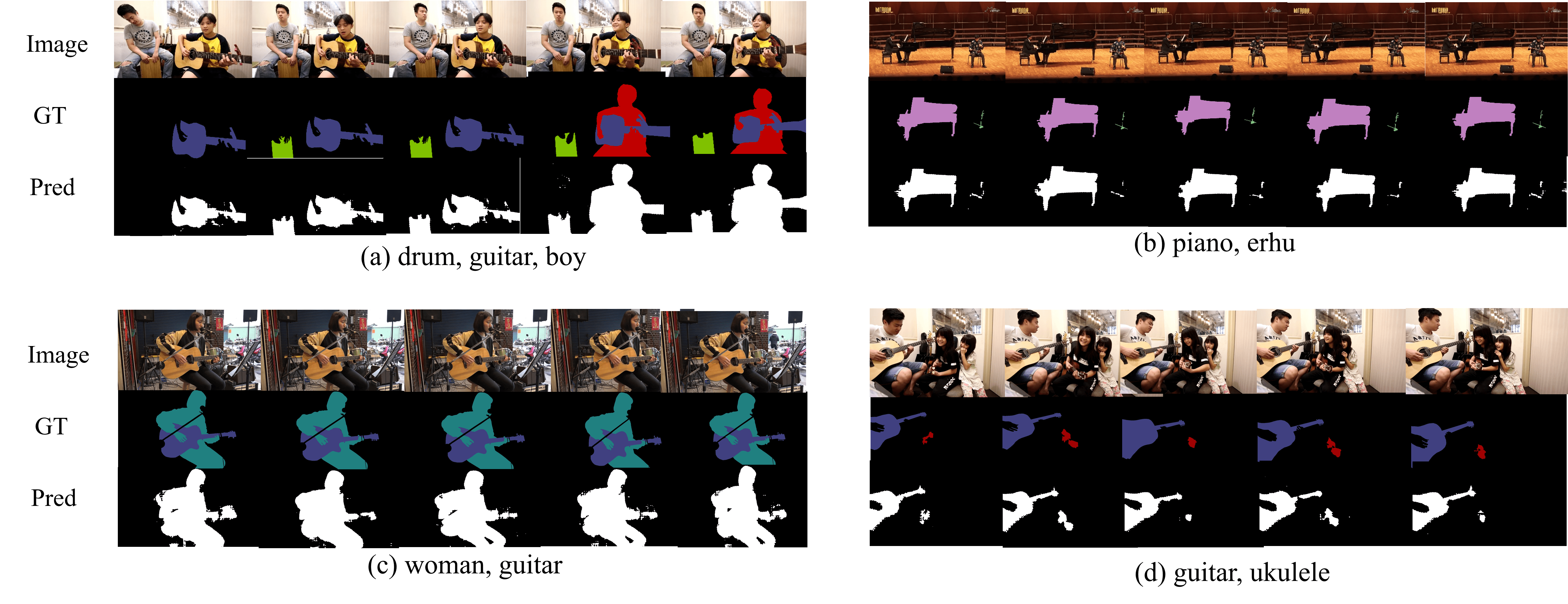}
     \caption{Visualization of segmentation results on the supervised AVS-V2 test set.}
     \label{fig:3_vis_ms}
\end{figure*}

\begin{figure*}[!htb]
     \centering     
     \includegraphics[width=1\textwidth]
     {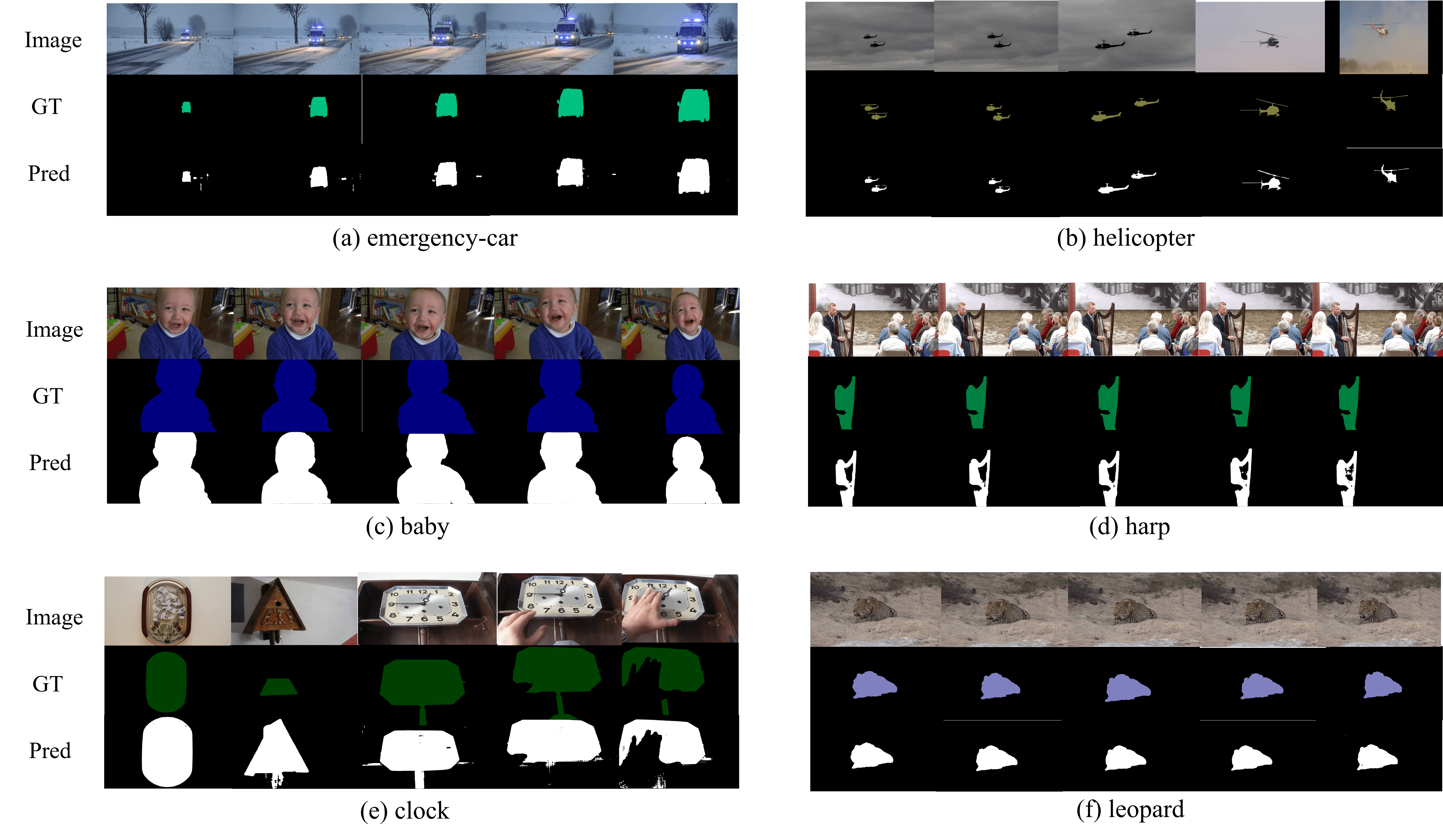}
     \caption{Visualization of segmentation results on the zero-shot AVS-V3 test set.}
     \label{fig:3_vis_v3}
\end{figure*}

\subsection{B.1 Grounding Segmentation Capability}
The effectiveness of the multi-source audio-visual segmentation capabilities can be evaluated from various qualitative perspectives. As shown in Figure \ref{fig:3_vis_v2}, we select three representative test results to demonstrate the effectiveness of our model, where the white areas represent the segmentation results. In Results (a), the model exhibits the ability to perform fine-grained segmentation. Despite the obstruction caused by the fingers on the keyboard, our model accurately excludes the finger regions while preserving the visible keyboard pixels between the finger gaps. Results (b) demonstrate the model's capability in multi-source audio segmentation, successfully separating Cajon, Guitar, and Human (voice) based on their respective sound sources in the image. Results (c) showcase the model's dynamic tracking ability, as it can segment a train moving closer from a distance based on both the audio and visual cues.

\subsection{B.2 Multi-source Segmentation}
 
In this work, a challenge is the multi-source AVS, because the model needs to analyze multiple sound sources from a mono-channel waveform and map them to the visual space. As illustrated in Figure \ref{fig:3_vis_ms}, we screen out some examples of multi-source AVS in visually unchanging environments, showing that the incorporation of audio input is necessary to aid the visual foundation model in accurately localizing sounding objects.

\subsection{B.3 Generalizable Zero-shot Segmentation}
We assess the model's generalization capability on unseen objects using the AVS-V3 dataset. As depicted in Figure \ref{fig:3_vis_v3}, our model achieves impressive segmentation results for these objects, despite the fact that their corresponding categories are not included in the training process. Notably, when examining the (e) ``clock'' group, we observe that our model could effectively extend its segmentation ability to different instances of the same unseen category. It is important to acknowledge that there are instances of slight pixel over-segmentation (e.g., in the second ``clock'') and under-segmentation (e.g., in the ``harp''). However, this is primarily attributed to the granularity of the manual segmentation process. Nevertheless, we can confidently assert that the model successfully segments the sounding objects.

\section*{C Versatility}
Our method \textit{works for various transformer-based architectures}, like Mask2Former (M2F). We switch the foundation model to M2F with Swin-S as the visual backbone, achieving better results (mIoU=59.69$|$F=0.757) with fewer params.

\section*{D Comparison with the most related methods}

We for the first time introduce the encoder-prompt-decoder paradigm in AVS, whereas other methods are fusion-based (Table \ref{tab:compare}). The most related is SAM-Fusion, which uses the same VFM. We also use audio to \textit{prompt visual prior knowledge}. Our GAVS is distinct from previous ones, and the results indicate its effectiveness. 

\begin{table}[!htbp]
\centering
\resizebox{0.8\columnwidth}{!}{
\begin{tabular}{lc}
\hline
Model & Methodology \\ \hline
AVSBench & Base: decode from fused audio-visual. \\
AVSegFormer & Base + audio as a query. \\ \hline
SAM-Fusion (ours) & Base + audio as a query. \\
Audio-SAM (ours) & Audio prompts visual w/o visual prior. \\
GAVS (ours) & Audio prompts visual w/ visual prior. \\ \hline
\end{tabular}
}
\caption{Methods comparison. Last three use SAM as VFM.}
\label{tab:compare}
\end{table}

\end{document}